\documentclass[letterpaper]{article}
\usepackage{proceed2e}
\usepackage[margin=1in]{geometry}

\usepackage{graphicx}
\graphicspath{{figures/}}
\DeclareGraphicsExtensions{.pdf,.jpeg,.png}

\usepackage{bm}

\usepackage[ruled,linesnumbered]{algorithm2e}

\usepackage{makecell}

\newcommand{\overbar}[1]{\mkern 1.5mu\overline{\mkern-1.5mu#1\mkern-1.5mu}\mkern 1.5mu}

\usepackage{multirow}

\usepackage{times}

\usepackage[round]{natbib}   
\title{A Concept Learning Tool Based On Calculating Version Space Cardinality}

\author{ 
{\bf Kuo-Kai Hsieh} \\
University of California, Santa Barbara \\
kuokai@umail.ucsb.edu \\
\And
{\bf Li-C. Wang} \\
University of California, Santa Barbara \\
licwang@ece.ucsb.edu 
}

\begin{document}

\maketitle

\begin{abstract}
In this paper, we proposed VeSC-CoL (Version Space Cardinality based Concept Learning) to deal with concept learning on extremely imbalanced datasets, especially when cross-validation is not a viable option. VeSC-CoL uses version space cardinality as a measure for model quality to replace cross-validation. Instead of naive enumeration of the version space, Ordered Binary Decision Diagram and Boolean Satisfiability are used to compute the version space. Experiments show that VeSC-CoL can accurately learn the target concept when computational resource is allowed. 
\end{abstract}

\section{INTRODUCTION}

In hardware Electronic Design Automation (EDA) and Test, learning methods have been widely used in diverse applications to analyze data from design simulation and testing of silicon chips \citep{wang2017experience}. In many applications, data can be encoded with discrete features and the underlying learning approach can be formulated as concept learning. However, in those applications, the applicability of concept learning is usually limited by the availability of the data. For example, the data comprises two classes of samples, positive and negative, and there are very few or no positive samples. This scenario occurs quite often when the goal of learning is to uncover the cause for some special property observed on the positive samples. To give an example, in functional verification (e.g. for a System-on-Chip (SoC) design) the data are simulation traces based on applying a set of input stimuli to the design. The interest can be on a particular property observed in the simulation traces. Usually, there are few traces showing the property. The learning goal is to understand what characteristics in the input stimuli lead to those few traces, i.e. the causes. In this context, features are used to describe the input stimuli. 

Abstractly, the learning problem faced in those applications can be formulated as the following. Two classes of samples are given, $D^+$ and $D^-$. Each sample $s$ is encoded with a set of discrete features $f_1, \ldots, f_n$. Let $m_p=|D^+|$ and $m_n=|D^-|$. Usually we have $m_p \ll m_n$, $m_p$ is very small or can be zero. The cause(s) of the positive samples can be described as a target concept. And the learning goal is to uncover this target concept from $D^+$ and $D^-$. Without loss of generality, we can assume the features are binary. This is because multiple binary features can be used to encode a single multi-value feature. Furthermore, we can assume the target concept falls into the scope of a $k$-term disjuctive normal form (DNF) with
a small $k$. In practice, a single cause for a property can be described as a combination of feature values, e.g. a monomial. If a property has multiple causes, then they can be described as a $k$-term DNF. Because the number of causes for a property is small (e.g. 2 to 4), usually there is no need to consider a large $k$.  

Cross-validation is a common way to select and validate a learning model in practice. However, for the concept learning problem described above, cross-validation is not a viable option due to the lack of positive samples. Because the samples are obtained through simulation or silicon measurement, there is a significant cost to get an additional sample. By the nature of the problem, getting more positive samples is difficult without knowing the cause, i.e. without learning the target concept. 

Because cross-validation cannot be used, in practice validation of a learning result relies on a separate process outside the learning. For example, this validation may involve expert's investigation of a target concept or through discussion among a group of engineers to determine its meaningfulness. Because the cost associated with such a model validation process can be significant, it is desirable for a learning tool to output a model with some sort of guarantee. 

In traditional concept learning, there is no requirement that the output concept is unique. In other words, the version space can still contain many concepts that can fit the data and the output concept is just one of them. Usually, a tool can report several models if the user adjusts some learning parameters. However, a tool does not report the size of the version space, i.e. how many remaining concepts that can all fit the data. 

Suppose we have a tool that can calculate the size of version space. Then, this tool can be used not only to find the concept that fits the data but also to identify the concept hypothesis space that results in a small version space based on the data. In other words, a learning strategy can be implemented to search for the hypothesis space assumption that fits the data where the "fitting" can be defined as resulting in a very small version space, e.g. size $<10$. Note that given a set of hypothesis space assumptions to choose from, it is possible that none of them can fit the data. In this case, the learning fails. Then, the user has two choices, either to expand the data or to include additional hypothesis space assumptions. 

Based on the discussion above, in this paper, we present Version Space Cardinality based Concept Learning (VeSC-CoL). The idea of VeSC-CoL is to search for a fitting hypothesis space with increased complexity. This implies that VeSC-CoL works with a set of pre-defined hypothesis space assumptions where the complexity of each hypothesis space is defined and can be calculated. Outputs from the VeSC-CoL tool include a learned concept and the version space cardinality as a quality measure for the concept. For example, if the cardinality is one, this means that the concept is unique and this indicates the highest quality for the model. If the cardinality is small, the tool can also list all fitting concepts. 

\section{VERSION SPACE LEARNING}

Version space learning originates in \citep{mitchell1978version}. Given a hypothesis space $H$, a set of positive samples $D^+$, and a set of negative samples $D^-$, version space is defined as the set of hypotheses that are consistent to the given samples. Formally speaking, $VS =$ $\{h \in H|\forall s \in D^+ \; h(s) = 1,  \forall s \in D^- \; h(s) = 0 \}$.

In \citep{mitchell1978version}, version space ($VS$) is represented by the boundary sets $S$ and $G$, where $S$ is the set of most specific hypotheses in $VS$ and $G$ is the set of most general hypotheses in $VS$. Then, for every hypothesis $h \in VS$, $h$ is as or more general than some hypothesis in $S$ and as or more specific than some hypothesis in $G$.

Following research in version space learning include investigating new methods to manipulate $S$ and $G$, e.g.  incremental version-space merging that utilizes set intersection idea to calculate the boundary sets \citep{hirsh1994generalizing}, proposing alternative version space representations \citep{lau2003programming}, and arguing that representation is not the key problem in version space learning \citep{hirsh2004version}.

Our work differs from the above research mainly in that we investigate methods to calculate version space cardinality. Except for naive enumeration, there is no published work solving this problem.

\section{VeSC-CoL OVERVIEW}

In this work, VeSC-CoL adopts a particular learning strategy enabled by the capability to calculate version space cardinality. Given a general assumption of hypothesis space, a sequence of hypothesis sub-spaces are defined. Each sub-space comprises hypotheses of the same complexity. VeSC-CoL then tries to find the simplest hypothesis sub-space that fits the data. 

\subsection{HYPOTHESIS SPACE PARTITION}

The search of VeSC-CoL can be thought of as following the Occam's razor principle to find the simplest hypothesis to fit the data as well. The added-value of VeSC-CoL is that it also identifies the simplest fitting hypothesis sub-space if it exists. As mentioned above, it is possible that a fitting hypothesis can be found while the simplest fitting hypothesis sub-space does not exist because the version space cardinality is too large. Hence, VeSC-CoL can fail. This is in contrast to a traditional learning tool where if there exists a fitting model, the tool would not fail. Take decision tree classifier as an example. Such learning can also follow the Occam's razor principle where the process of node splitting stops when a node contains only one class of data. 
Such learning would not fail if there exists a decision tree to fit the data. 

With above strategy, the learning problem can be stated as the following: given a seqeunce of hypothesis sub-spaces $H_1, H_2, \ldots$, a complexity measure $comp$, a hypothesis fitting evaluator $fit$, a cardinality bound $B$, and data $D$, find a hypothesis $h \in H_i$ such that $comp(h)$ is minimized subject to (1) $fit(h, D)$ is true and (2) the cardinality of the version space based on $H_i$ is $\leq B$. 

\begin{figure}[h]
\centering
\includegraphics[width=0.45\textwidth]{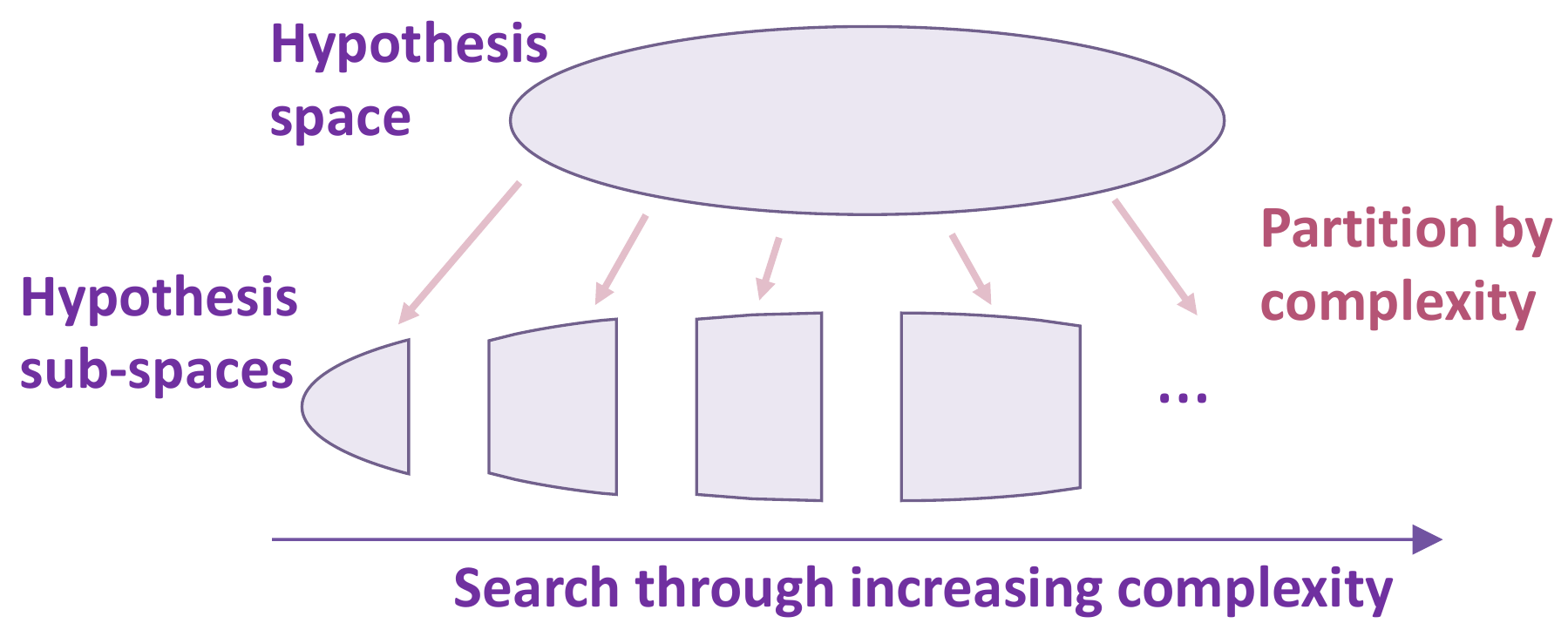}
\caption{Sub-space partitioning based on complexity}
\label{fig:minimization_by_search}
\end{figure}

Figure \ref{fig:minimization_by_search} illustrates a simple search process to solve the learning problem. 
Given the complexity measure $comp$, the original hypothesis space $H$ is partitioned into a sequence of
hypothesis sub-space $H_1, H_2, \ldots$ where each sub-space comprises the hypotheses of the same complexity. 
The search proceeds from the lowest-complexity sub-space to highest-complexity sub-space. The search process stops when it first finds a hypothesis that satisfies the two constraints.

Suppose $H_i$ is a hypothesis sub-space where every hypothesis $h$ does not satisfy the first constraint
($fit(h, D)$ is not true). This indicates an under-fitting situation. On the other hand, suppose 
$H_j$ for $j>i$ contains a hypothesis $h$ satisfying the first constraint but the version space of
$H_j$ violates the second constraint. This indicates an over-fitting situation. 
If $j=i+1$, then there is no solution to the learning problem. This means that there is no hypothesis sub-space to fit the data or there is not enough data to obtain a small enough version space.

Recall that in this work VeSC-CoL is applied for learning a $k$-term DNF formula. Given a $k$-term formula $h$, the complexity measure $comp$ used by VeSC-CoL is defined as $comp(h) = (l, k)$, where $k$ is the number of terms and $l$ is the number of literals in $h$. We say $(l_1, k_1) < (l_2, k_2)$ if and only if $l_1 < l_2$ or $l_1 = l_2 \;\land\; k_1 < k_2$.

For example, suppose our hypothesis space is $k$-term DNF with $k$ no greater than $3$. The partitioned sub-space in the order of increasing complexity is $(1,1), (2,1), (2,2), (3,1), (3,2), (3,3), (4,1), \dots$.

\subsection{VeSC-CoL FLOW}

\begin{figure}[h]
\centering
\includegraphics[width=0.45\textwidth]{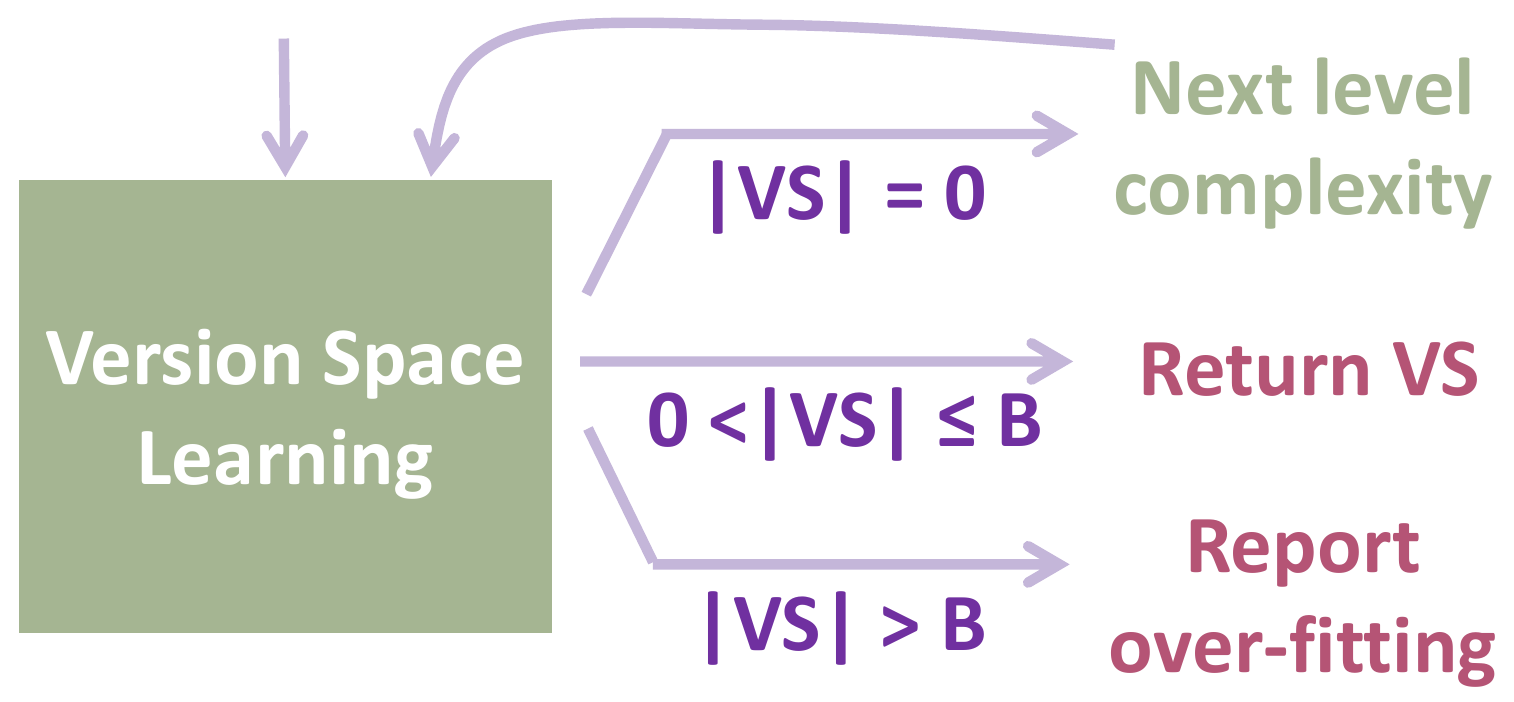}
\caption{Illstration of VeSC-CoL flow}
\label{fig:flow}
\end{figure}

Figure \ref{fig:flow} depicts the flow of VeSC-CoL. It starts from calculating the version space cardinality of the simplest-complexity sub-space. If there is no hypothesis consistent with the data, i.e. $|VS| = 0$, then VeSC-CoL moves onto the next hypothesis sub-space. The iteration stops when both constraints are satisfied. At this point, VeSC-CoL reports at most $B$ hypotheses in version space as well as version space cardinality as a measure of learning quality. $B$ is an application-specific parameter and usually is set to the maximum number of hypotheses that a user can handle in model evaluation.

In theory, this problem is proved to be no easier than \#P-complete \citep{hirsh2004version}. Even determining whether a version space is empty or not for $k$-term DNF is NP-hard \citep{pitt1988computational}.

Though this problem is intractable in theory, in practice a useful tool can still be developed. In this paper, we propose two methods to calculate version space cardinality. The first method is based on Ordered Binary Decision Diagram (BDD) \citep{bryant1986graph}. The version space cardinality can be obtained by calculating the number of minterms in BDD. The second method is based on Boolean Satisfiability (SAT). For the SAT implementation, VeSC-CoL does not calculate version space cardinality. Rather, VeSC-CoL tries to find at most $B+1$ hypotheses in the version space. 

\section{BDD-BASED LEARNING}

\begin{figure}[h]
\centering
\includegraphics[width=0.4\textwidth]{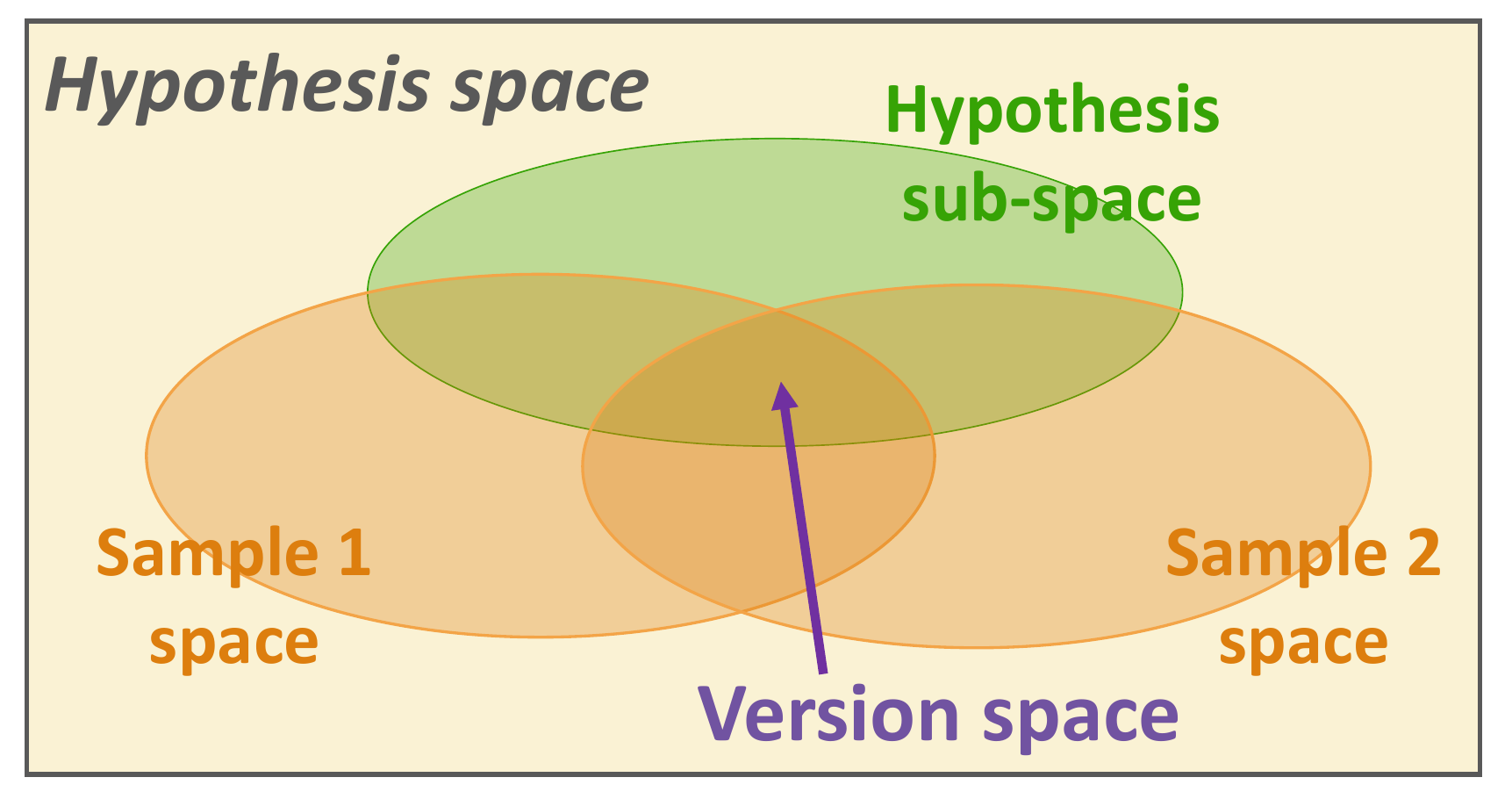}
\caption{Version space learning by set intersection}
\label{fig:version_space_learning}
\end{figure}

The basic idea of the proposed BDD-based version space learning method is based on set intersection \citep{hirsh1994generalizing}. Figure \ref{fig:version_space_learning} illustrates this idea. BDD is used to represent a set of hypothesis. In a version space BDD, there is a bijection between a minterm of the BDD and a DNF representation.

First, a set of hypotheses with a given complexity is created, which is the hypothesis sub-space in Figure \ref{fig:version_space_learning}. Then each sample is converted into the set of hypotheses that agree with the sample. The version space can be obtained by intersecting the hypothesis sub-space and all the sample spaces.

Set intersection can be performed via Boolean AND operation of BDDs. To determine the size of version space, because of bijection, we can simply count the number of minterms in the version space BDD. Note that if a hypothesis has multiple representations, special treatment is required and it is discussed in Section \ref{sec:non-canonicality}.

In the description below, we assume the hypothesis sub-space is a $(l,k)$-space, i.e. a $k$-term $l$-literal DNF. If another hypothesis sub-space definition is used, the BDD encoding method will be different but the encoding ideas can be reused.

\subsection{IDEA OF ENCODING}
\begin{figure}[h]
\centering
\includegraphics[width=0.35\textwidth]{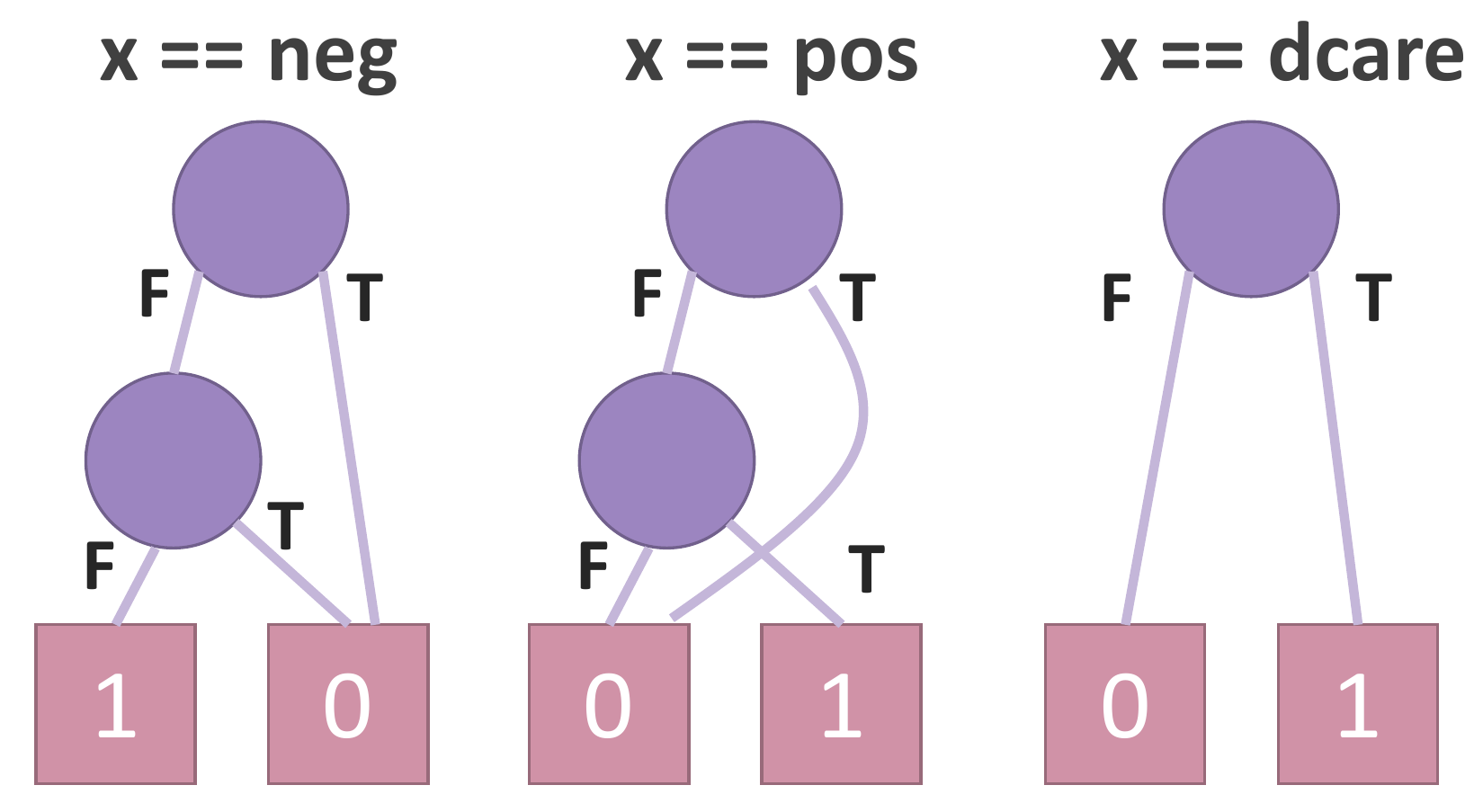}
\caption{Using BDD to encode a three-value variable}
\label{fig:bdd_encoding_idea}
\end{figure}

Given the number of features $n$ and the number of terms $k$. Let $x^j_i$ represent the status of the $i$-th feature in the $j$-th term, wherein $x^j_i \in$ \{neg, pos, dcare\}, which denotes appearing in negative form, in positive form and don't care (not appearing). Since $x^j_i$ is a three-value variable, we use two Boolean variables to represent it in BDD. An example is shown in Figure~\ref{fig:bdd_encoding_idea}. In sum, there are $2nk$ variables in BDD.

\subsection{BASE HYPOTHESIS SPACE ENCODING}
Algorithm \ref{alg:hypo_bdd_base} shows the method to create a BDD representing an $n$-feature and $k$-term DNF hypothesis space. This algorithm simply forces each $x^j_i$ to be in its three possible values. Note that there is a bijection between a satisfiable assignment in the returned BDD and a $k$-term DNF representation.

\begin{algorithm}[ht]
\label{alg:hypo_bdd_base}
\SetAlgoLined
\KwIn{Integers $n$, $k$}
\KwOut{BDD $dd$}
dd $\leftarrow$ BDD\_One()\;
\For{ $i \in \{1, 2, \dots , n\}$}{
    tmp\_dd = BDD\_Or($x^j_i$ == $pos$, $x^j_i$ == $neg$, $x^j_i$ == $dcare$)\;
    dd $\leftarrow$ BDD\_And(dd, tmp\_dd)\;
}
\Return $dd$\;
\caption{Creating a BDD representing a $n$-variable, $k$-term DNF space}
\end{algorithm}

\subsection{HYPOTHESIS SUB-SPACE ENCODING}
Algorithm \ref{alg:hypo_bdd1} describes the method to create a BDD representing an $n$-feature, $k$-term DNF, $l$-literal space. Line 1 to Line 4 is the initialization process for dynamic programming. The idea is that at the end, an assignment that makes lit\_dd[$w$] == 1 can be mapped to a DNF formula having at least $w$ literals.

\begin{algorithm}[ht]
\label{alg:hypo_bdd1}
\SetAlgoLined
\KwIn{Integers $n$, $k$, $l$}
\KwOut{BDD $dd$}
lit\_dd[0] $\leftarrow$ BDD\_One()\;
\For{ $w \in \{1, 2, \dots , l+1\}$}{
    lit\_dd[$w$] $\leftarrow$ BDD\_Zero()\;
}

\For{ $j \in \{1, 2, \dots , k\}$}{
    \For{ $i \in \{1, 2, \dots , n\}$}{
    var\_dd $\leftarrow$ BDD\_Or($x^j_i$ == $pos$, $x^j_i$ == $neg$)\;
        \For{ $w \in \{l+1, l, \dots , 1\}$}{
        tmp\_dd $\leftarrow$ BDD\_And(lit\_dd[$w-1$], var\_dd)\;
        lit\_dd[$w$] $\leftarrow$ BDD\_Or(lit\_dd[$w$],  tmp\_dd)\;
        }
    }
}
$dd$ $\leftarrow$ BDD\_And(lit\_dd[$l$], BDD\_Not(lit\_dd[$l+1$]))\;
\Return $dd$\;
\caption{Creating a BDD representing a $n$-variable, $k$-term, $l$-literal space}
\end{algorithm}

Line 5 to Line 13 updates the BDDs used in dynamic programming.
lit\_dd is used to save intermediate results. At line 10, an assignment that makes lit\_dd[$w$] == 1 can be mapped to a DNF formula with at least $w$ literals in the processed $x^j_i$. The outer two loops iterate all $x^j_i$, then for each $x^j_i$, lit\_dd is updated accordingly. Note that lit\_dd is updated from the highest index to the lowest index. After processing all the $x^j_i$, line 14 creates the result BDD representing the $l$-literal sub-space. 

Note that the returned BDD does not guarantee a bijection between a minterm and a DNF representation due to the NOT operation. To have this bijection, a minterm in the returned BDD must be in the base hypothesis space BDD as well. Also, with proper BDD variable ordering, it can be shown that the complexity of Algorithm \ref{alg:hypo_bdd1} is $O(nkl)$.

\subsection{POSITIVE SAMPLE SPACE ENCODING}
Algorithm \ref{alg:pos_bdd} converts a positive sample to a BDD representing a set of consistent hypotheses. Its input parameters are $n$, the number of features, $k$, the number of terms, and $s[i] \in \{0, 1\}$, the value of the $i$-th feature. The key idea is in line 12, given a hypothesis, at least one term of the hypothesis must be evaluated as true so the hypothesis is evaluated as true.

\begin{algorithm}[h]
\label{alg:pos_bdd}
\SetAlgoLined
\KwIn{Integers $n$, $k$, and an n-dimensional Boolean vector $s$}
\KwOut{BDD $dd$}

$dd$ $\leftarrow$ BDD\_Zero()\;
\For{ $j \in \{1, 2, \dots , k\}$}{
    term\_dd $\leftarrow$ BDD\_One()\;
    \For{ $i \in \{1, 2, \dots , n\}$}{
        \uIf{$s[i]$ == 0}{
        tmp\_dd $\leftarrow$ BDD\_Or($x^j_i$ == $neg$, $x^j_i$ == $dcare$)\;
        }
        \Else{
        tmp\_dd $\leftarrow$ BDD\_Or($x^j_i$ == $pos$, $x^j_i$ == $dcare$)\;
        }
    term\_dd $\leftarrow$ BDD\_And(term\_dd, tmp\_dd)\;
    }
    $dd$ $\leftarrow$ BDD\_Or($dd$, term\_dd)\;
}

\Return $dd$\;
\caption{Converting a positive sample to BDD}
\end{algorithm}

Suppose $s = 101$. For a single term to be evaluated as true, feature $1$ and feature $3$ must not be negative literals and feature $2$ must not be a positive literal in the term. Otherwise, this term is evaluated as false. The generalization of this idea shown in line 3 to line 11. 

At line 10, each minterm in term\_dd can be mapped to a single DNF term. At line 14, each minterm in $dd$ can be mapped to a $k$-term DNF formula. With proper BDD variable ordering, the complexity of this algorithm \ref{alg:pos_bdd} is $O(kn)$.

\subsection{NEGATIVE SAMPLE SPACE ENCODING}

The algorithm of converting a negative sample to its space BDD is similar to algorithm \ref{alg:pos_bdd}. The differences are (1) all the terms must be evaluated as false and (2) the conversion rule for a single term is negated. Algorithm \ref{alg:neg_bdd} shows the conversion algorithm.

\begin{algorithm}[h]
\label{alg:neg_bdd}
\SetAlgoLined
\KwIn{Integers $n$, $k$, and an n-dimensional Boolean vector $s$}
\KwOut{BDD $dd$}

$dd$ $\leftarrow$ BDD\_One()\;
\For{ $j \in \{1, 2, \dots , k\}$}{
    term\_dd $\leftarrow$ BDD\_Zero()\;
    \For{ $i \in \{1, 2, \dots , n\}$}{
        \uIf{$s[i]$ == 0}{
        tmp\_dd $\leftarrow$ ($x^j_i$ == $pos$)\;
        }
        \Else{
        tmp\_dd $\leftarrow$ ($x^j_i$ == $neg$)\;
        }
    term\_dd $\leftarrow$ BDD\_Or(term\_dd, tmp\_dd)\;
    }
    $dd$ $\leftarrow$ BDD\_And($dd$, term\_dd)\;
}

\Return $dd$\;
\caption{Converting a negative sample to BDD}
\end{algorithm}

Again, suppose $s = 101$. For a single term to be evaluated as false, one of the following conditions must hold: at least one of feature $1$ and feature $3$ appears as a negative literal, or feature $2$ appears as a positive literal. The generalization of this idea shown in line 3 to line 11. At line 12, since all the terms must be evaluated as false, the result $dd$ is the AND of all the term\_dd. With proper BDD variable ordering, the complexity of this algorithm \ref{alg:neg_bdd} is $O(kn)$.

\subsection{OBTAINING VERSION SPACE}

Version space can be obtained by performing an AND of all the above BDDs. Recall that each BDD represents a set of hypotheses inside the hypothesis sub-space, which agree with a positive sample or a negative sample. The AND is equivalent to the set intersection operation so the resulting BDD represents the version space.

In actual implementation, the AND of a set of BDDs is accomplished by performing a sequence of AND operations on two BDDs. 
We observed that the ordering of AND operations on BDDs significantly influences the runtime. There can be two prferences: (1) Process the hypothesis sub-space BDD first and (2) If $k \leq2 $, process positive sample BDDs before negative sample BBDs; otherwise process negative sample BDDs before positive sample BDDs.

\begin{figure}[h]
\centering
\includegraphics[width=0.45\textwidth]{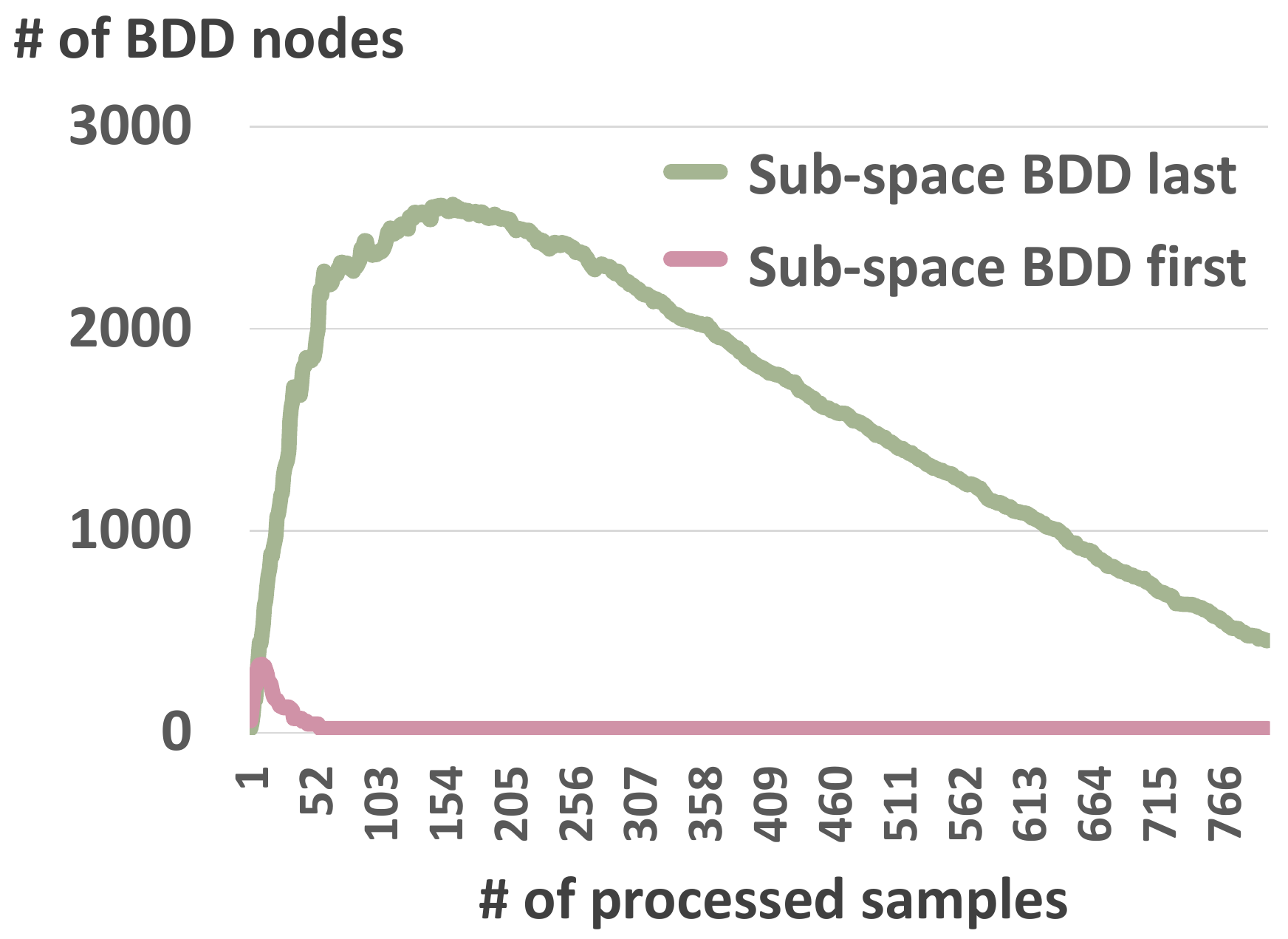}
\caption{Example to illustrate that processing hypothesis sub-space BDD at the beginning 
is more efficient}
\label{fig:BDD_constraint_ordering}
\end{figure}

To illustrate the first preference, Figure \ref{fig:BDD_constraint_ordering} shows the number of BDD nodes in the version space BDD versus the number of processed samples. There are 100 features, 3 positive samples, and 800 negative samples. For the red line, the first AND operation is applied to the hypothesis sub-space BDD and a positive sample BDD. The next two ANDs involve the remaining two positive sample BDDs. The negative sample BBDs are processed afterward. For the green line, the first three positive sample BDDs are processed first, followed by processing negative sample BDDs. The hypothesis sub-space BDD is processed last. 

The runtime is proportional to the number of BDD nodes. It can be clearly observed that the difference in runtime between the two cases is significant. The reason is that processing the hypothesis sub-space BDD first can more effectively trim the version space.

Table \ref{tbl:BDD_pos_neg_ordering} shows the runtime comparison between 
processing positive sample BDDs first and processing negative sample BDDs first. 
The comparison is presented as a ratio between the two. 
In each case, there are 100 features, 250 positive samples, and 250 negative negative samples. 
The number of literals $l$ is randomly selected in each run and $l \leq 15$. 
In each case, there are 10 runs for the positive first and 10 runs for the negative first.
A geometric mean of the 10 runtimes is calculated. Then, the ratio is calculated
from the two geometric means. The reason to use the geometric mean is that
the 10 runtimes can differ significantly based on the selection of $l$. 
Table \ref{tbl:BDD_pos_neg_ordering} shows that for $k \leq2 $, processing positive sample BDDs 
before negative sample BDDs saves time, and vice versa.

\begin{table}[h]
\vspace{-0.3cm}
\caption{Runtime ratio of processing positive sample BDDs first over processing negative sample BDDs first}
\label{tbl:BDD_pos_neg_ordering}
\begin{center}
\begin{tabular}{cc}
\textbf{k}  & \textbf{pos-first/neg-first} \\
\hline 
1 & $1.89 * 10^{-6}$\\
2 & $8.70 * 10^{-2}$\\
3 & $1.76 * 10^{3}$ \\ 
4 & $2.38 * 10^{5}$ 
\end{tabular}
\end{center}
\vspace{-0.2cm}
\end{table}

For $k=1$, the problem is monomial learning. For monomial learning, it is well known that
positive samples are far more important than negative samples, i.e. positive
samples are far more effective to reduce the version space than negative samples. 
As a result, processing positive sample BDDs first is more effective. 
This property seems to somewhat carry over to the case $k=2$. It is interesting
that the situation reverses for $k=3$ and $k=4$. The theoretical reason for this
reverse is still unclear and should be investigated further in the future.

\subsubsection{Handling Non-Canonicality}
\label{sec:non-canonicality}
A hypothesis can be represented by different DNF formulas, e.g. $a + b = b + a$. Hence the size of version space cannot be obtained by counting the number of minterms in the version space BDD in general. Here we introduce another BDD that forces each term in a DNF representation to be in lexicographical order, which reduces the permutation among terms. Algorithm \ref{alg:lexi_order_bdd} shows the procedure to create a BDD having lexicographical order among two terms. In total $k-1$ such BDDs are required. Next, when the number of minterms in the version space BDD is in the same order as $B$, we convert each minterm to its DNF formula and then use a BDD to represent it. Since BDD is a canonical representation, we are able to obtain the size of version space.

\begin{algorithm}[h]
\label{alg:lexi_order_bdd}
\SetAlgoLined
\KwIn{Integers $n$, $k_1$, $k_2$}
\KwOut{BDD $dd$}

$dd$ $\leftarrow$ BDD\_Zero()\;
eq\_dd $\leftarrow$ BDD\_One()\;
\For{ $i \in \{1, 2, \dots , n\}$}{
    cond1 $\leftarrow$ BDD\_And($x^{k_1}_i == neg$, $x^{k_2}_i == pos$)\;
    cond2 $\leftarrow$ BDD\_And($x^{k_1}_i == neg$, $x^{k_2}_i == dcare$)\;
    cond3 $\leftarrow$ BDD\_And($x^{k_1}_i == pos$, $x^{k_2}_i == dcare$)\;
    cond = BDD\_Or(cond1, cond2, cond3)\;
    tmp\_dd $\leftarrow$ BDD\_And(eq\_dd, cond)\;
    $dd$ $\leftarrow$ BDD\_Or($dd$, tmp\_dd)\;
    eq\_dd $\leftarrow$ BDD\_And(eq\_dd, $x^{k_1}_i == x^{k_2}_i$)\;
}

\Return $dd$\;
\caption{BDD representing the $k_1$-th term is lexicographically smaller than the $k_2$-th term}
\end{algorithm}

\section{SAT-BASED LEARNING}
The idea of SAT-based version space learning is similar to BDD-based encoding. The basic components are the same: the hypothesis sub-space, the positive sample spaces, and the negative sample spaces. The main difference is that SAT requires others encoding techniques to restrict the number of clauses and the number of symbols. 
Let $n$ be the number of features, $l$ be the number of literals, $k$ be the number of terms, $m_p$ be the number of positive samples and $m_n$ be the number of negative samples.
Then, the encoding method described below results in a CNF formula with $\Theta(nkl)$ symbols and $\Theta(nkl + nkm_p + km_n)$ clauses.

Same as BDD encoding, each feature can appear in positive, negative or does not appear in a term. Hence, three symbols are used to represent each case.
\begin{itemize}
\item $X^j_{i,1}$ is True iff the $i$-th feature in the $j$-th term appears in negative form
\item $X^j_{i,2}$ is True iff the $i$-th feature in the $j$-th term appears in positive form
\item $X^j_{i,3}$ is True iff the $i$-th feature in the $j$-th term does not appear 
\end{itemize}

Since exactly one of the three cases is true, one-hot constraints are required to enforce the requirement:

\begin{center}
$\Pi_{i=j}^k\Pi_{i=i}^n (X^j_{i,1}+X^j_{i,2}+X^j_{i,3})(\lnot X^j_{i,1}+ \lnot X^j_{i,2})$ \\
$(\lnot X^j_{i,1}+ \lnot X^j_{i,3})(\lnot X^j_{i,2}+ \lnot X^j_{i,3})$.
\end{center}

\begin{figure*}[t!]
\centering
    \begin{minipage}[]{.45\textwidth}
        \includegraphics[width=0.9\textwidth]{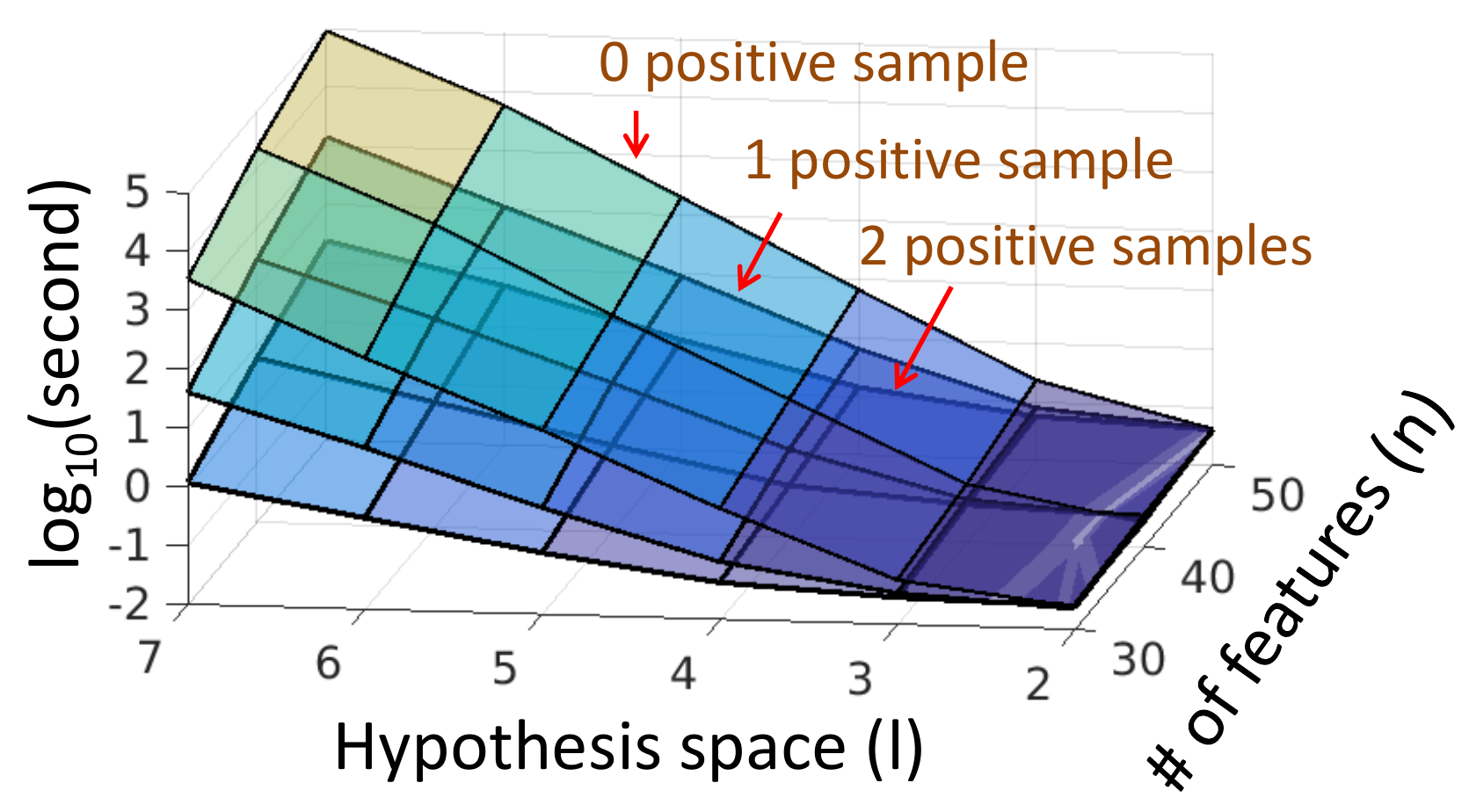}
        \caption{BDD runtime}\label{fig:bdd_runtime}
    \end{minipage}\qquad
    \begin{minipage}[]{.45\textwidth}
        \includegraphics[width=1.15\textwidth]{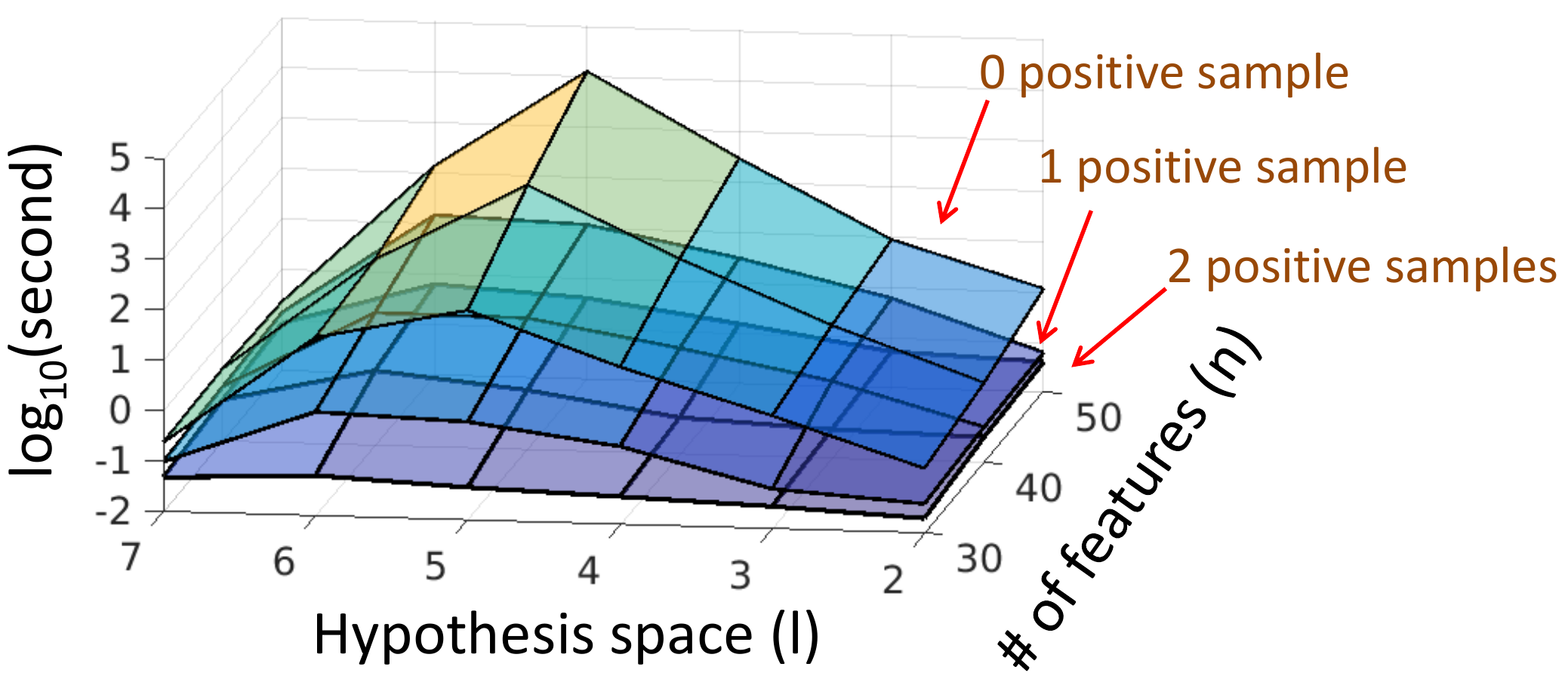}
        \caption{SAT runtime}\label{fig:sat_runtime}
    \end{minipage}
\end{figure*}

\subsection{HYPOTHESIS SUB-SPACE ENCODING}
For a given $(l, k)$, we need to constrain the space to contain only $l$ literals. It is the cardinality constraint.
The performance of different encoding methods for a cardinality constraint can be found in \citep{Ben-Haim2012}. 
In our implementation, we choose the sequential counter method \citep{Sinz2005} because its performance is comparable to other encoding methods and it has the unit propagation property \citep{Ben-Haim2012}.
The encoding formula shown in \citep{Sinz2005} cannot be used directly because it is for cardinality $\leq l$. A straightforward modification is used for cardinality $= l$, based on converting the sequential counter circuit to SAT clauses. The encoding for the cardinality constraint requires additional $l(nk-1)$ new symbols and $\Theta(nkl)$ clauses.

For this encoding, we use the same notation and symbol in \citep{Sinz2005}, so it is easier to get the difference between the modification and the original encoding, wherein $k$ is the number of symbols passing to the cardinality constraint and $S_{i,j}$ are additional symbols.

\begin{center}
$(x_1 + \lnot S_{1,1}) (\lnot x_1 + S_{1,1}) $, \\
$\Pi_{j=2}^{k} (\lnot S_{1,j})$, \\
$\Pi_{i=2}^{n} (\lnot x_i + \lnot S_{i-1, k})$, \\
$\Pi_{i=2}^{n-1} (x_i + S_{i-1, 1} + \lnot S_{i, 1})(\lnot x_i + S_{i, 1})(\lnot S_{i-1, 1} + S_{i, 1})$, \\
$
\Pi_{i=2}^{n-1}\Pi_{j=2}^{k} (x_i + S_{i-1, j} + \lnot S_{i, j})(S_{i-1, j-1}  + S_{i-1, j} + \lnot S_{i, j}) (\lnot S_{i-1, j} + S_{i, j})(\lnot x_i + \lnot S_{i-1, j-1} + S_{i, j}) 
$,  \\
$(x_n + S_{n-1,k}) ( S_{n-1,k-1} + S_{n-1,k}) $. \\
\end{center}

\subsection{POSITIVE SAMPLE SPACE ENCODING}
Again, given a positive sample $s = 101$. For a single term to be evaluated as true, feature $1$ and feature $3$ must not appear in negative form and feature $2$ must not appear in positive form. Then, at least one term must be evaluated as true. A naive encoding leads to $n^k$ clauses, which is not feasible. To overcome this challenge, additional $k$ symbols, $A^1$, $A^2$, $\dots$, $A^k$, are used such that $A^j$ is true if and only if the $j$-th term is evaluated as true. With these additional symbols, the number of clauses reduces to $(n+1)k+1$. The requirement of at least one term is evaluated as true is encoded by a single clause:

\begin{center}
$(\Sigma_{j=1}^{k} A^j)$,
\end{center}

and for each $j$, the relation of $A^j$ and $X^j_{i,\delta}$ is maintained by 
\begin{center}
$\Pi_{i=1}^{n} (\lnot X^j_{i, 2-s[i]} + \lnot A^j)$, and \\
$(\Sigma_{i=1}^{n} X^j_{i, 2-s[i]} + A^j)$.
\end{center}

\subsection{NEGATIVE SAMPLE SPACE ENCODING}
Given a negative sample $s = 101$. For a single term to be evaluated as false, at least one of feature $1$ and feature $3$ must appear in negative form or feature $2$ appear in positive form. Besides, all the terms must be evaluated as false. For each sample, $k$ clauses are required and each clause encodes that a term is evaluated as false. The overall encoding is 
\begin{center}
$\Pi_{j=1}^{k} (\Sigma_{i=1}^{n} X^j_{i, 2-s[i]})$.
\end{center}

\subsection{SIZE OF VERSION SPACE}
Each satisfiable assignment in the above SAT problem can be mapped to a DNF formula. The size of version space can be obtained by counting the number of satisfiable assignments. A common approach is to add new clauses to remove previous satisfiable assignments and then call the SAT solver again. Removing a satisfiable assignment is a standard approach and omitted here. Note that we use the same approach to deal with the non-canonicality problem described in BDD-based learning, except the lexicographical order constraint is represented by SAT clauses.

\section{EXPERIMENTS}

We use CUDD-3.0.0 \citep{somenzi2015cudd} to implement the BDD-based learning and use Lingeling \citep{SAT2013} for the SAT-based learning. The dynamic variable re-ordering option in CUDD is disabled to facilitate the study of various aspects of the tool performance.

\subsection{RUNTIME COMPARISON}

We observed different characteristics of runtime between the SAT-based method and the BDD-based method. 
For example, we use a simple experiment to illustrate their differences. 
In this experiment, the target concept is assumed to be a 5-literal monomial (i.e. $k=1$ and $l=5$). There are 1000 randomly generated negative samples. There can be 0, 1, and 2 positive samples. 
Figure \ref{fig:sat_runtime} and Figure \ref{fig:bdd_runtime} show the runtime results. 
Each point is the average of runtimes over 10 runs. The size bound $B$ is set to $1$, so if the size of version space is large than $1$, the SAT-based learning would stop after finding the second fitting hypothesis. Note that in this experiment, for all cases with $l<5$ the calculated size of version space is always $0$, for $l=5$ the calculated size of version space is exactly $1$, and for $l>5$ the calculated size of version space is always larger than $1$. This shows that both learning methods can identify the correct hypothesis sub-space and the correct hypothesis. 

\begin{figure}[h]
\centering
\includegraphics[width=0.4\textwidth]{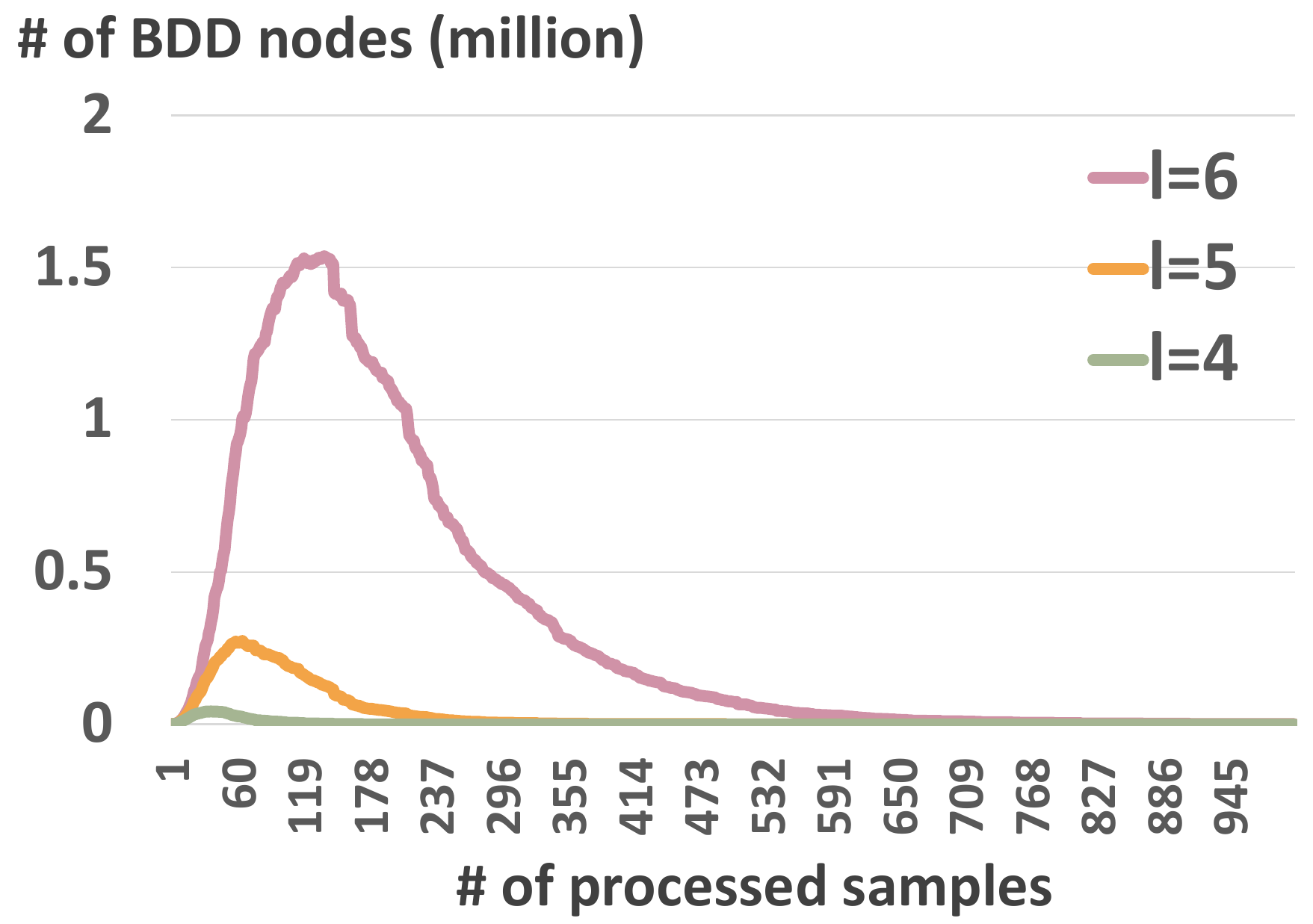}
\vspace{-10pt}
\caption{The peak number of nodes grows as $l$ increases}
\label{fig:bdd_peaks}
\end{figure}

The results show that the runtime of the BDD-based method is exponential to $l$. On the other hand, 
the SAT-based method has a peak runtime at $l=5$, i.e. the size of version space is $1$. 
Figure \ref{fig:bdd_peaks} shows another interesting property of BDD-based learning where
the number of positive sample is 1. 
In each case, the positive sample BDD is processed first, followed by processing the
negative sample BDDs. The figure shows the number of BDD nodes as a function of the
number of processed samples. As it can be observed, the peak number of BDD nodes occurs
earlier in the process than later, for example within the first 200 samples. This implies
that the computational limitation occurs within the processing of the first 200 samples. 
As a result, it is not the case that a larger dataset implies a longer run time. 
As mentioned above, the deciding factor for the runtime is the length $l$.

\begin{figure}[h]
\centering
\includegraphics[width=0.4\textwidth]{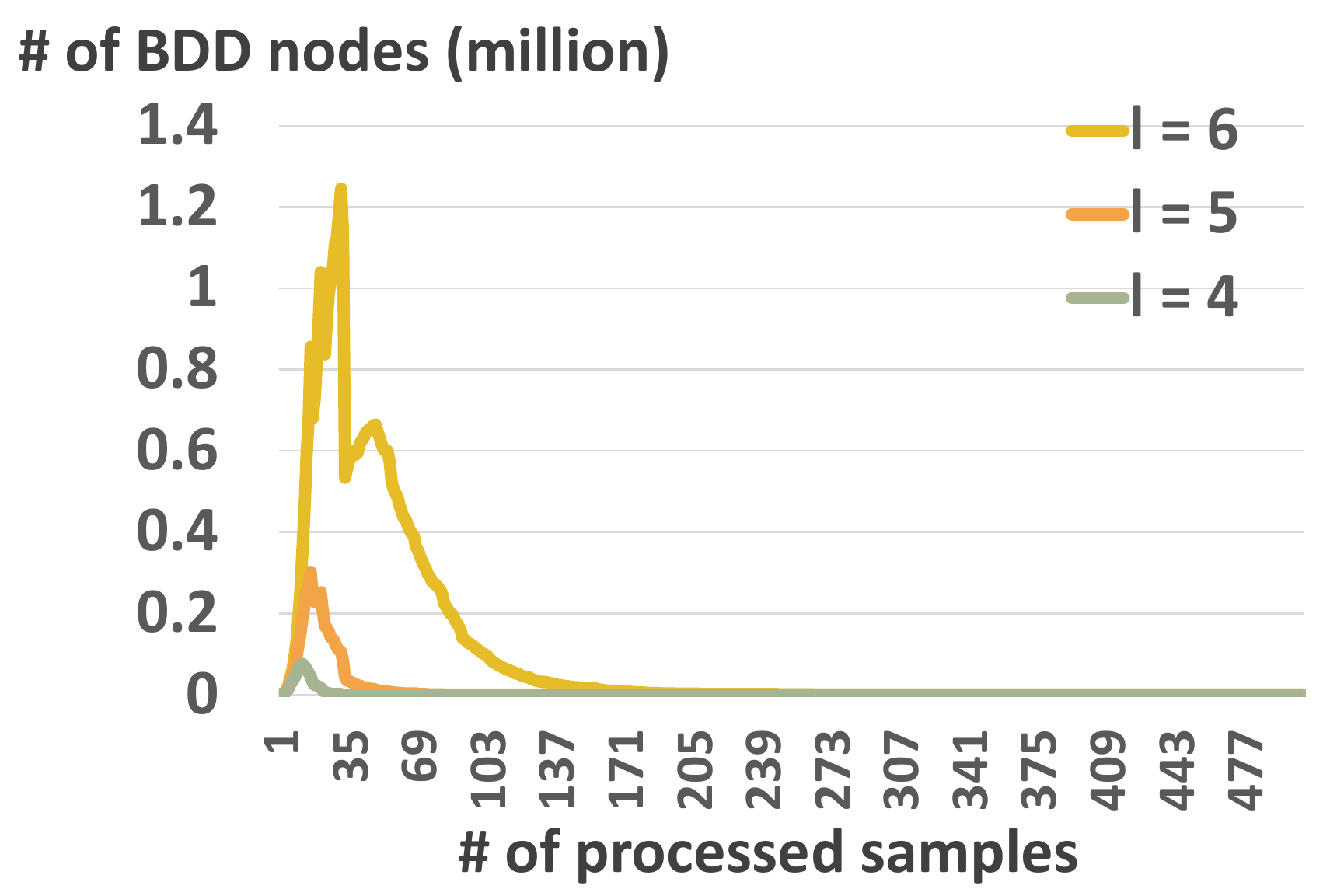}
\vspace{-10pt}
\caption{Similar result for 2-term DNF}
\label{fig:bdd_peaks_2}
\end{figure}

Figure~\ref{fig:bdd_peaks_2} shows similar runtime results as those shown in Figure \ref{fig:bdd_peaks}.
In this experiment, the target concept is a 2-term DNF with $l=5$ where one term is of length 2
and the other item is of length 3. 
The number of positive samples is 5, the number of negative samples is 500,
and the number of features is 100. Similarly, positive sample BDDs are processed first. 
Observe that the peak number of BDD nodes also occurs earlier in the process and the length $l$ is the deciding factor for the runtime. 

\begin{table*}[h]
\caption{VeSC-CoL learns the target monomial, while CART and ID3 produce irrelevant results}
\label{tbl:useless_learning_result}
\begin{center}
\begin{tabular}{ccc}
\textbf{VeSC-CoL}  & \textbf{CART} & \textbf{ID3} \\
\hline 
$x_2x_{63}\overbar{x_{75}}x_{78}\overbar{x_{80}}$ & 
$x_3x_4x_{28}x_{47}\overbar{x_{53}}\overbar{x_{55}}\bm{\overbar{x_{80}}}$ & 
$\bm{x_2}x_3x_4\overbar{x_{30}}x_{47}\overbar{x_{53}}\overbar{x_{81}}$
\\
$x_{39}\overbar{x_{45}}x_{72}\overbar{x_{74}}x_{95}$ &
$\overbar{x_{5}}x_{16}x_{35}\bm{\overbar{x_{45}}}\overbar{x_{55}}\overbar{x_{56}}x_{59}$ &
$x_8x_{40}\bm{\overbar{x_{45}}}x_{64}\bm{\overbar{x_{74}}}x_{87}$ 
\\
$\overbar{x_{2}}\overbar{x_{14}}x_{52}\overbar{x_{57}}x_{87}$ &
$x_{11}\bm{\overbar{x_{14}}}\overbar{x_{24}}x_{61}x_{64}x_{90}\overbar{x_{92}}$ &
$\overbar{x_{5}}\overbar{x_{6}}x_{16}x_{35}\overbar{x_{45}}\overbar{x_{56}}x_{59}$
\\
$x_{40}\overbar{x_{45}}x_{64}\overbar{x_{74}}x_{87}$ &
$\overbar{x_{4}}x_{8}\bm{\overbar{x_{45}}}\overbar{x_{47}} \bm{x_{64}}\bm{\overbar{x_{74}}}\overbar{x_{89}}$ &
$\overbar{x_{2}}\overbar{x_{14}}\overbar{x_{24}}x_{61}\bm{x_{64}}x_{90}\overbar{x_{92}}$
\\
$\overbar{x_{57}}x_{58}x_{77}\overbar{x_{95}}x_{98}$ &
$\overbar{x_{5}}x_{29}x_{38}\overbar{x_{43}}\overbar{x_{79}}x_{99} + \overbar{x_{3}}\overbar{x_{5}}\overbar{x_{29}}x_{38}\overbar{x_{43}}x_{49}\overbar{x_{79}}x_{99}$ &
$\overbar{x_{5}}x_{6}\overbar{x_{11}}\overbar{x_{14}}\overbar{x_{18}}\overbar{x_{34}}x_{45}$
\end{tabular}
\end{center}
\vspace{-10pt}
\end{table*}

\subsection{COMPARISON WITH OTHER METHODS}

To compare VeSC-CoL with other concept learning methods such as 
CART \citep{cart84} and ID3 \citep{quinlan1986induction}, we continue the experiment above
where the target concept is a 5-literal monomial. There are 100 features, 2 positive samples, 
and 1000 negative samples. 
Table \ref{tbl:useless_learning_result} shows the learning result. In each case, VeSC-CoL is able to correctly
identify the target concept. On the other hand, the results from CART are less meaningful. For the first three tasks, each CART result has only 1 literal relevant to the target concept while providing 6 unrelated literals and missing 4 literals in the target. For the last task, the CART learning result is a 2-term DNF in which no feature is related to the target. The learning results from ID3 are dissimilar to the target concept as well.



\subsection{COMPLEXITY ORDERING}

As mentioned before, for two hypothesis sub-space $H_i$ and $H_j$ of $k$-term DNF,
we consider the complexity of $H_i$ is smaller than $H_j$ if $l_i<l_j$ where
$l_i$ and $l_j$ are the numbers of literals in the hypotheses in $H_i$ and $H_j$, respectively. 
Recall that each hypothesis sub-space comprises hypotheses of equal length. 
Note that this complexity ordering is based on two main reasons: (1) As shown above,
BDD-based learning is sensitive to the length $l$. Hence, the ordering ensures that
the learning processes the computationally-easier hypothesis sub-spaces first. 
(2) In practice, a concept with a smaller length is easier to interpret than that
with a larger length. Therefore, it is preferred to uncover a shorter concept if possible. 

\subsection{ACCURACY OF VeSC-CoL} 

In the experiments to compare VeSC-CoL with CART and ID3, we observe that VeSC-CoL
can always uncover the correct answer. Note that it is possible to construct a dataset
to fool the VeSC-CoL tool so that it reports an incorrect answer even with the cardinality
bound $B=1$. However, with randomly generated datasets, we observe that when the data
is sufficiently large and $B=1$, VeSC-CoL 
always finds the correct target concept assuming the concept is in the hypothesis
space considered (e.g. 3-term DNF up to length 15). 
In particular, we observed in the following experiments that VeSC-CoL
always find the correct answer:

\begin{itemize}
\vspace{-3pt}
\item All 1-term DNF cases with up to 100 features and $l$ up to 7. The number of positive 
samples can be 0 to 2 and the number of negative samples is 10000. 

\item All 1-term DNF cases with up to 500 features and $l$ up to 8. The number of positive samples is larger than 5 and the number of negative samples is 10000. 

\item All 2-term DNF cases with up to 100 feature and $l$ up to 8. For each term, there exists a positive sample that can be explained only by the term. The number of positive samples is larger than 5 and the number of negative samples is 10000. 

\item All 3-term DNF cases with up to 100 feature and $l$ up to 9. For each term, there exists a positive sample that can be explained only by the term. The number of positive samples is larger than 10 and the number of negative samples is 10000. 

\end{itemize}

\section{CONCLUSION AND FUTURE WORK}

We propose VeSC-CoL, a version space cardinality based concept learning tool, for
learning extremely imbalanced datasets. We use experiment results to note several
key properties of the tool. VeSC-CoL is applicable without cross-validation. 
The version space cardinality bound is used to control the quality of the learning
result. In our study, we observed that VeSC-CoL can always identify the correct
target concept assuming that the concept is included in one of the hypothesis
sub-spaces to be analyzed. VeSC-CoL is supported by two implementations, one based
on BDD and the other based on SAT. Their runtimes can be quite different. Therefore VeSC-CoL runs the two methods in parallel and stops when one of them completes.

One challenge is to generalize the encoding method. The current encoding method is closely related to the $k$-term DNF representation and the complexity measure. Suppose a hypothesis is represented in BDD and the complexity measure is the number of BDD nodes, the encoding will be different. Given a hypothesis representation and a complexity measure, finding an encoding method is a non-trivial task that needs further investigation.

While the experiments show several interesting properties of the implementations, 
further research is required to analyze the theoretical reasons behind those properties and
to formalize their descriptions. While the current work focuses on the development of 
the tool, its performance in actual applications (such as those in EDA and Test) will be 
assessed and reported in the near future. 





\bibliography{uai18}

\begin{thebibliography}{13}
\providecommand{\natexlab}[1]{#1}
\providecommand{\url}[1]{\texttt{#1}}
\expandafter\ifx\csname urlstyle\endcsname\relax
  \providecommand{\doi}[1]{doi: #1}\else
  \providecommand{\doi}{doi: \begingroup \urlstyle{rm}\Url}\fi

\bibitem[Ben-Haim et~al.(2012)Ben-Haim, Ivrii, Margalit, and
  Matsliah]{Ben-Haim2012}
Yael Ben-Haim, Alexander Ivrii, Oded Margalit, and Arie Matsliah.
\newblock Perfect hashing and {CNF} encodings of cardinality constraints.
\newblock In \emph{International Conference on Theory and Applications of
  Satisfiability Testing}, pages 397--409. Springer, 2012.

\bibitem[Biere(2013)]{SAT2013}
Armin Biere.
\newblock Lingeling, plingeling and treengeling entering the sat competition
  2013.
\newblock \emph{Proceedings of SAT Competition}, 2013.

\bibitem[Breiman et~al.(1984)Breiman, Friedman, Olshen, and Stone]{cart84}
L.~Breiman, J.~Friedman, R.~Olshen, and C.~Stone.
\newblock \emph{{Classification and Regression Trees}}.
\newblock Wadsworth and Brooks, Monterey, CA, 1984.

\bibitem[Bryant(1986)]{bryant1986graph}
Randal~E Bryant.
\newblock Graph-based algorithms for boolean function manipulation.
\newblock \emph{Computers, IEEE Transactions on}, 100\penalty0 (8):\penalty0
  677--691, 1986.

\bibitem[Hirsh(1994)]{hirsh1994generalizing}
Haym Hirsh.
\newblock Generalizing version spaces.
\newblock \emph{Machine Learning}, 17\penalty0 (1):\penalty0 5--46, 1994.

\bibitem[Hirsh et~al.(2004)Hirsh, Mishra, and Pitt]{hirsh2004version}
Haym Hirsh, Nina Mishra, and Leonard Pitt.
\newblock Version spaces and the consistency problem.
\newblock \emph{Artificial Intelligence}, 156\penalty0 (2):\penalty0 115--138,
  2004.

\bibitem[Lau et~al.(2003)Lau, Wolfman, Domingos, and Weld]{lau2003programming}
Tessa Lau, Steven~A Wolfman, Pedro Domingos, and Daniel~S Weld.
\newblock Programming by demonstration using version space algebra.
\newblock \emph{Machine Learning}, 53\penalty0 (1-2):\penalty0 111--156, 2003.

\bibitem[Mitchell(1978)]{mitchell1978version}
Tom~Michael Mitchell.
\newblock Version spaces: an approach to concept learning.
\newblock Technical report, STANFORD UNIV CALIF DEPT OF COMPUTER SCIENCE, 1978.

\bibitem[Pitt and Valiant(1988)]{pitt1988computational}
Leonard Pitt and Leslie~G Valiant.
\newblock Computational limitations on learning from examples.
\newblock \emph{Journal of the ACM (JACM)}, 35\penalty0 (4):\penalty0 965--984,
  1988.

\bibitem[Quinlan(1986)]{quinlan1986induction}
J.~Ross Quinlan.
\newblock Induction of decision trees.
\newblock \emph{Machine learning}, 1\penalty0 (1):\penalty0 81--106, 1986.

\bibitem[Sinz(2005)]{Sinz2005}
Carsten Sinz.
\newblock Towards an optimal {CNF} encoding of boolean cardinality constraints.
\newblock In \emph{International Conference on Principles and Practice of
  Constraint Programming}, pages 827--831. Springer, 2005.

\bibitem[Somenzi(2015)]{somenzi2015cudd}
Fabio Somenzi.
\newblock {CUDD}: {CU} decision diagram package release 3.0.0. {P}ublic
  software, {U}niversity of {C}olorado at {B}oulder.
\newblock 2015.

\bibitem[Wang(2017)]{wang2017experience}
Li-C Wang.
\newblock Experience of data analytics in eda and test—principles, promises,
  and challenges.
\newblock \emph{IEEE Transactions on Computer-Aided Design of Integrated
  Circuits and Systems}, 36\penalty0 (6):\penalty0 885--898, 2017.

\end{thebibliography}

\end{document}